\documentclass[runningheads]{llncs}
\usepackage[T1]{fontenc}
\usepackage{graphicx}
\usepackage{color}
\usepackage{xcolor}
\usepackage{amsmath, amssymb}
\usepackage[inline]{enumitem}
\usepackage{algorithm}
\usepackage{algpseudocode}
\usepackage{booktabs}

\usepackage[dvipsnames]{xcolor}
\usepackage{tikz}
\usepackage{pgfplots}
\usepackage{graphicx} 
\usepackage{caption}  
\usepackage{subcaption}
\usepackage{multirow}
\usepackage{wrapfig}
\usepackage{array}

\newcommand{\best}[1]{\textcolor{red}{#1}}
\newcommand{\second}[1]{\textcolor{Green}{#1}}
\newcommand{\third}[1]{\textcolor{blue}{#1}}

\usepackage{tabularx}
\usepackage{array}
\newcolumntype{Y}{>{\centering\arraybackslash}X}

\begin{document}
\title{Optimizing In-Context Demonstrations for LLM-based Automated Grading}
%
\author{Yucheng Chu \inst{1} \and Hang Li \inst{1} \and Kaiqi Yang \inst{1} \and Yasemin Copur-Gencturk \inst{2} \and Kevin Haudek \inst{1} \and Joseph Krajcik \inst{1} \and Jiliang Tang \inst{1}}
\institute{Michigan State University  \and University of Southern California	}
\authorrunning{Y. Chu et al.}
%
\maketitle              
\begin{abstract}
Automated assessment of open-ended student responses is a critical capability for scaling personalized feedback in education. While large language models (LLMs) have shown promise in grading tasks via in-context learning (ICL), their reliability is heavily dependent on the selection of few-shot exemplars and the construction of high-quality rationales. Standard retrieval methods typically select examples based on semantic similarity, which often fails to capture subtle decision boundaries required for rubric adherence. Furthermore, manually crafting the expert rationales needed to guide these models can be a significant bottleneck. To address these limitations, we introduce \textsc{GUIDE} (\textbf{G}rading \textbf{U}sing \textbf{I}teratively \textbf{D}esigned \textbf{E}xemplars), a framework that reframes exemplar selection and refinement in automated grading as a boundary-focused optimization problem. \textsc{GUIDE} operates on a continuous loop of selection and refinement, employing novel contrastive operators to identify ``boundary pairs'' that are semantically similar but possess different grades. We enhance exemplars by generating discriminative rationales that explicitly articulate why a response receives a specific score to the exclusion of adjacent grades. Extensive experiments across datasets in physics, chemistry, and pedagogical content knowledge demonstrate that \textsc{GUIDE} significantly outperforms standard retrieval baselines. By focusing the model's attention on the precise edges of rubric, our approach shows exceptionally robust gains on borderline cases and improved rubric adherence. 
\textsc{GUIDE} paves the way for trusted, scalable assessment systems that align closely with human pedagogical standards.

\end{abstract}

\vspace{-0.3in}
\section{Introduction}
\label{sec:introduction}
\vspace{-0.1in}

Automated assessment is a central goal in artificial intelligence in education (AIED), offering the potential for timely feedback and scalable support in large, diverse classrooms \cite{burrows2015eras}.
Within this domain, automatically grading open-ended short answers remains technically demanding. Unlike keyword matching, valid assessment requires interpreting semantic nuances, evaluating causal reasoning and handling the immense linguistic variability students use to express identical concepts.
Traditional supervised models, which rely on task-specific features and large annotated datasets often struggle to transfer effectively across changing curricula or distinct student populations due to overfitting \cite{burrows2015eras}. 
Large language models (LLMs) have transformed this landscape by enabling \emph{in-context learning} (ICL), allowing models to perform complex assessments given only a rubric and labeled demonstrations \cite{brown2020language,liu2023pre}.
This paradigm is particularly appealing for educational assessments as it leverages the model's pre-trained world knowledge to generalize across different topics and question types without extensive, task-specific feature engineering. 
Despite these advantages, the reliability of LLM-based in-context graders is often compromised by their sensitivity to the selection and ordering of exemplars
\cite{zhao2021calibrate,min2022rethinking}. This sensitivity is exacerbated by the noise inherent in authentic classroom data, where student responses exhibit various writing styles and partial reasoning that are difficult to separate \cite{burrows2015eras,horbach2019influence}. This in part reflects developmental learning, as learners take up new ideas and gradually develop more sophisticated understanding of disciplinary topics.

To systematically enhance reliability, we approach the problem through two complementary perspectives: enriching the \emph{informational content} of individual exemplars and optimizing the \emph{discriminative composition} of the exemplar set. 
The first perspective addresses the ambiguity inherent in mapping noisy student text to a score. Augmenting examples with high-quality chain-of-thought (CoT) rationales can significantly improve model performance by explicitly breaking down the grading process \cite{wei2022chain}. In educational contexts, expert-written rationales are particularly valuable, as they not only justify the score but also model the specific pedagogical tone and rubric adherence required for valid assessment. However, manual curation of such expert explanations is labor-intensive and often prohibitively expensive to scale. To address this bottleneck, we seek to synthesize these high-quality rationales. By pairing existing responses with generated reasoning, we can construct robust exemplars without sole reliance on human experts.

The second pathway focuses on defining the precise decision boundaries within the context window. We posit that the reliability issue often stems from suboptimal context composition, where the selected examples fail to clarify edge cases. Standard selection strategies, such as similarity-based retrieval \cite{liu2021makes}, often fail in this domain because they prioritize semantic proximity over pedagogical utility.
In a grading context, retrieving an example that is semantically similar to the student's response is insufficient. 
Effective grading prompts require not merely representative examples, but \emph{contrastive pairs} that appear semantically similar but receive different scores. This forces the model to attend to the nuanced differences that define the grade.
While research in other domains has attempted to enhance ICL \cite{wan2025few,rubin2022learning}, they have significant limitations when applied to automated grading.
For example, BRIDGE \cite{wan2025few} treats exemplars as independent discrete units and optimizes a set to maximize global accuracy. While effective for tasks like mathematical problem solving where answers are definitive, this approach often fails to capture subtle distinctions required for rubric adherence in educational settings. 
These challenges raise a pivotal question regarding how we can automatically construct and refine in-context exemplars that attend to boundary cases to maximize grading reliability under realistic constraints.

To address this challenge, we present \textbf{GUIDE} (\textbf{G}rading \textbf{U}sing \textbf{I}teratively \textbf{D}esigned \textbf{E}xemplars), an automated framework that reframes exemplar optimization in grading as a boundary definition problem. Unlike prior work that relies solely on global accuracy signals to select independent demonstrations, \textsc{GUIDE} enhances optimization process with the focus on \emph{discriminative interaction} between examples. The framework operates on an iterative loop of ``selection'' and ``refinement''. First, to \textcolor{YellowOrange}{\textbf{select}} the most effective set of exemplars, we introduce
novel contrastive operators within a constrained Bayesian optimization framework. Instead of randomly mutating the exemplar set, the contrastive operators actively seek boundary pairs that are semantically similar but possess different grades. The operators attempt to swap or insert these boundary cases into context window, forcing the optimizer to select exemplars that clearly delineates the decision boundary rather than just illustrating typical cases.
Since selecting boundary examples is insufficient if the model does not understand \emph{why} they differ, we adopt \emph{discriminative rationale \textcolor{Emerald}{\textbf{generation}}}. 
We prompt the LLM to generate contrastive rationales that explicitly articulate why a response receives a specific score \emph{and not} its adjacent neighbors (for instance, explaining why a response is a score of 1 rather than 0 or 2). 
By automating the construction of these pedagogical boundary pairs, \textsc{GUIDE} not merely shows \emph{what} the typical examples are, but teaches \emph{where} the precise boundaries between distinct score levels are drawn. The optimized boundary pairs and their rationales are used as seeds to generate new synthetic rationales and select the ideal boundary cases for the next iteration. Finally, this optimal exemplar set with synthetic rationales can clarify the decision space while significantly reduce manual effort. We validate our approach through extensive empirical analysis on diverse educational datasets spanning the domains of physics science, chemistry, and pedagogical knowledge. Our research demonstrates that \textsc{GUIDE} consistently outperforms baselines by prioritizing the resolution of borderline cases, thereby offering a more robust and pedagogically valid foundation for automated grading.

\vspace{-0.1in}
\section{Related Work}
\label{sec:related_work}
\vspace{-0.1in}

\subsection{Automated Grading of Open-Ended Responses}
\vspace{-0.1in}
Automated grading for short-answer and essay scoring has evolved significantly over the past decades. Early systems relied heavily on hand-crafted features, which are fed into traditional supervised classifiers \cite{burrows2015eras}.
The deep learning era shifted the paradigm toward pre-trained language models like BERT and RoBERTa, which achieved state-of-the-art performance by fine-tuning on labeled datasets \cite{devlin2019bert,liu2019roberta}.
Most recently, the field has adopted LLMs via ICL, where the model grades responses given only a rubric and a few examples \cite{brown2020language}. While ICL reduces the labeling burden, recent studies highlight that off-the-shelf LLMs can be inconsistent, struggling with specific rubric constraints or exhibiting unsatisfactory performance based on student writing styles \cite{mizumoto2023exploring}.

\vspace{-0.1in}
\subsection{Prompt and Exemplar Optimization in ICL}
\vspace{-0.1in}

In-context learning (ICL) performance is highly sensitive to the formatting and ordering of demonstrations, often requiring careful engineering to maximize model accuracy \cite{zhao2021calibrate}. Dominant selection heuristics typically retrieve examples from a fixed pool based on semantic similarity or distribution coverage \cite{liu2021makes,su2022selective}. Similarity metrics in grading often fail to distinguish between correct reasoning and plausible-sounding misconceptions that share high lexical overlap \cite{rubin2022learning}. Furthermore, relying on a static pool prevents the correction of noisy or unclear student responses~\cite{burrows2015eras}. To address the fragility of fixed retrieval, recent research has shifted toward automating the design of instructions and demonstrations. Techniques like APE and the ``optimize-generate'' paradigm treat the prompt as a variable to be optimized, iteratively synthesizing or refining exemplars to improve performance \cite{zhou2022large,wan2025few}.

\vspace{-0.1in}
\section{Problem Statement}
\label{sec:problem_statement}
\vspace{-0.1in}
We formulate automated grading as a classification task that maps a student response $x \in \mathcal{X}$ to a discrete score $y \in \mathcal{Y}$ (e.g., $\mathcal{Y} = \{0, 1, 2\}$). We utilize a frozen large language model $\mathcal{M}$ to perform in-context learning, where predictions are conditioned on a context window composed of a task instruction $\mathcal{I}$, a grading rubric $\mathcal{R}$, and a set of demonstrations $\mathcal{E} = \{e_1, \dots, e_k\}$. Each demonstration is a tuple $e_j = (x_j, y_j, r_j)$, containing a reference response, its ground-truth score, and a rationale explaining the grading logic. The model predicts the score for a query response $x$ according to:
\begin{equation*}
    \hat{y} = \operatorname*{argmax}_{y \in \mathcal{Y}} P_{\mathcal{M}}(y \mid x, \mathcal{I}, \mathcal{R}, \mathcal{E})
\end{equation*}

In practice, we assume access to a training dataset $\mathcal{S}= \{(x_i, y_i)\}_{i=1}^N$. Optionally, a small subset of training examples $\mathcal{S}_{\text{exp}} \subseteq \mathcal{S}$ is associated with expert-annotated rationales (where $\mathcal{S}_{\text{exp}}$ may be $\emptyset$), while the remainder lacks reasoning annotations. Our objective is to optimize the demonstration set $\mathcal{E}$ for grading performance on a held-out validation set $\mathcal{V}$. Specifically, we seek to select a subset of training examples and generate effective rationales for them, forming an optimal demonstration set $\mathcal{E}^* = \{(x_j, y_j, r_j)\}_{j=1}^k$ where $(x_j, y_j) \in \mathcal{S}$, to maximize the prediction accuracy of $\mathcal{M}$.

\section{Method}
\label{sec:methodology}
\vspace{-0.1in}

\begin{figure}[htbp]
\vspace{-0.1in}
    \centering
    \includegraphics[width=1.05\linewidth]{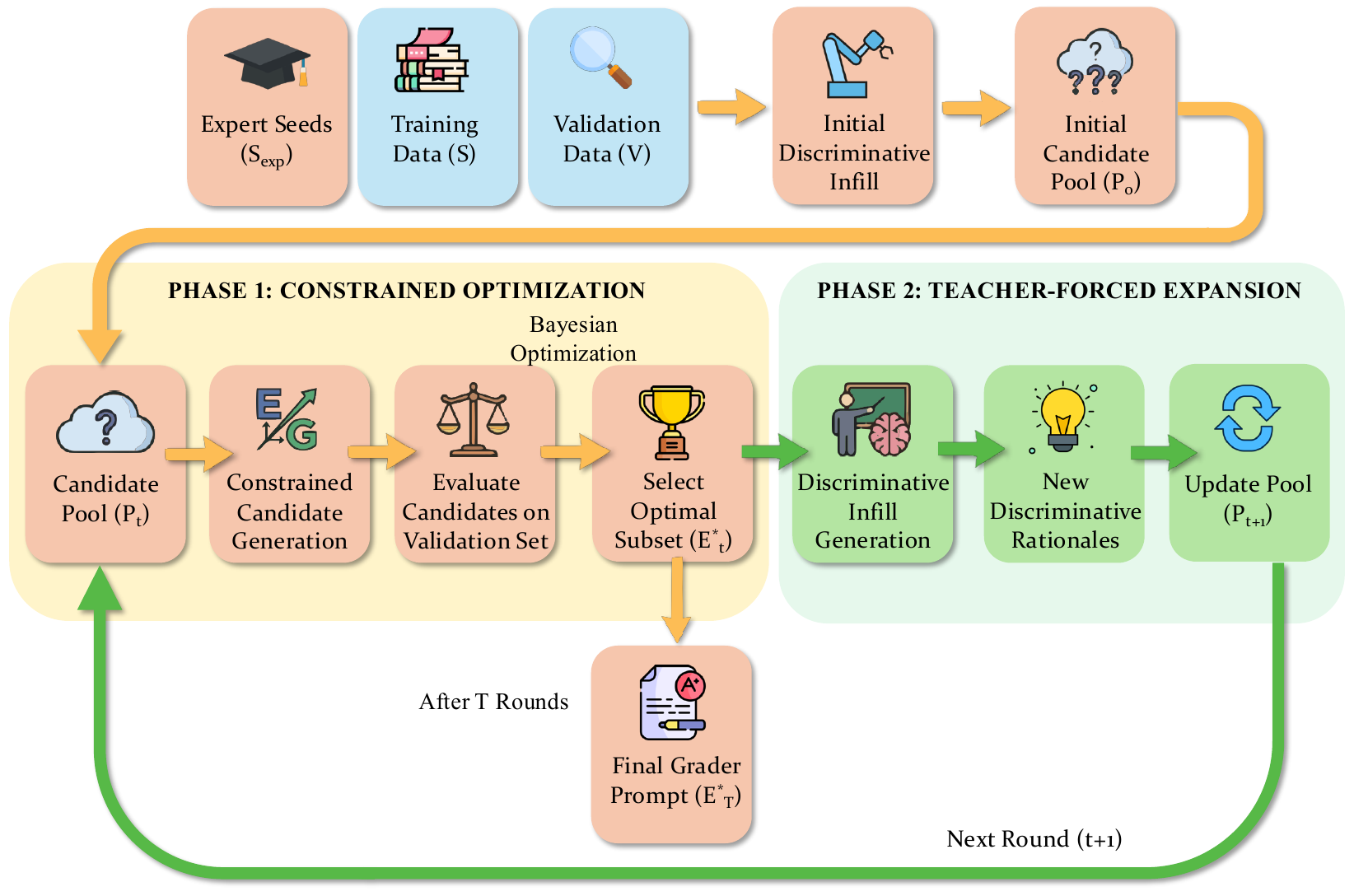}
    \vspace{-0.2in}
    \caption{An overview of the GUIDE algorithm.}
    \label{fig:framework}
\end{figure}

We propose \textbf{GUIDE} (\textbf{G}rading \textbf{U}sing \textbf{I}teratively \textbf{D}esigned \textbf{E}xemplars), an \textit{iterative} framework that constructs optimal in-context demonstration sets by targeting decision boundaries where nuanced grading errors occur. 
As illustrated in Fig.~\ref{fig:framework} and Algorithm~\ref{alg:guide}, GUIDE alternates between two phases over $T$ rounds. In the \textbf{\textcolor{YellowOrange}{selection}} phase, constrained Bayesian optimization identifies a boundary-focused demonstration set $\mathcal{E}^*$ from a candidate pool $\mathcal{P}$. In the \textbf{\textcolor{Emerald}{generation}} phase, discriminative rationale generation synthesizes rubric-compliant rationales via contrastive infilling, thereby enriching $\mathcal{P}$ for the subsequent iteration.


\subsection{Phase 0: Candidate Pool Initialization}
GUIDE operates over a dynamic \textbf{candidate pool} $\mathcal{P}$, which augments the training set $\mathcal{S}$ with generated rationales. Each exemplar in the pool is a triplet $e = (x, y, r)$, where $r$ is a rationale explaining why input $x$ warrants label $y$. Unlike
the static training set, the pool evolves across rounds as new rationales are
generated.
We initialize the pool by retaining expert exemplars with their human-authored rationales (if available) and generating initial rationales for all remaining training examples (Algorithm \ref{alg:guide} Line 1):
$\mathcal{P}_0 \leftarrow \mathcal{P}_{\text{exp}} \cup \{(x_i, y_i, r_i^{(0)}) : (x_i, y_i) \in \mathcal{S} \setminus \mathcal{S}_{\text{exp}}\}$
where $\mathcal{P}_{\text{exp}} = \{(x_i, y_i, r_i^{\text{exp}})\}_{(x_i, y_i) \in \mathcal{S}_{\text{exp}}}$ denotes the expert exemplars with their human-authored rationales, and $r_i^{(0)}$ is generated by the LLM conditioned on the expert exemplars as context.

\vspace{-8pt}

\subsection{Phase 1: Bayesian Optimization with Contrastive Selection}

Given the pool $\mathcal{P}_t$ at round $t$, we seek an optimal subset $\mathcal{E}_t^* \subseteq \mathcal{P}_t$ that, when used as the few-shot context, yields the highest alignment with the ground truth rubric. Direct optimization is computationally intractable due to the exponential search space of potential subsets and the high cost of executing an LLM inference pass for every evaluation. To address this, we employ Bayesian optimization (BO) to efficiently navigate the search space (Lines 4--18). This approach maintains a probabilistic surrogate model to approximate the mapping between an exemplar subset and its grading quality, allowing us to identify promising candidates with minimal computational expense.

\begin{algorithm}[t]
\caption{GUIDE Framework}
\label{alg:guide}
\begin{algorithmic}[1]
\Require $\mathcal{S}$, $\mathcal{S}_{\text{exp}}$, $\mathcal{V}$, rounds $T$, budget $n_{\text{eval}}$
\Ensure $\mathcal{E}^*_T$

\State Initialize $\mathcal{P}_0 \gets \mathcal{P}_{\text{exp}} \cup \{(x_i, y_i, r_i^{(0)})\}_{(x_i,y_i) \in \mathcal{S} \setminus \mathcal{S}_{\text{exp}}}$

\For{$t = 1$ to $T$}
    \State \textbf{Phase 1: \textcolor{YellowOrange}{Selection}}
    \State $\mathcal{D} \gets \emptyset$ \Comment{Observation history}
    \For{$i = 1$ to $n_{\text{init}}$} \Comment{GP Initialization}
        \State $\mathcal{E} \gets \textsc{RandomSubset}(\mathcal{P}_t)$
        \State $\mathcal{D} \gets \mathcal{D} \cup \{(\mathcal{E}, \textsc{Evaluate}(\mathcal{E}, \mathcal{V}))\}$
    \EndFor
    \State Fit GP on $\mathcal{D}$
    
    \For{$j = 1$ to $n_{\text{eval}} - n_{\text{init}}$} \Comment{BO Iterations}
        \State Generate candidates $\mathcal{C}$ via contrastive operators
        \State $\mathcal{E}_{\text{next}} \gets \arg\max_{\mathcal{E} \in \mathcal{C}} \text{EI}(\text{GP}, \mathcal{E})$
        \State $\mathcal{D} \gets \mathcal{D} \cup \{(\mathcal{E}_{\text{next}}, \textsc{Evaluate}(\mathcal{E}_{\text{next}}, \mathcal{V}))\}$
        \State Refit GP on $\mathcal{D}$
    \EndFor
    
    \State $a^* \gets \max_{(\mathcal{E}, \cdot) \in \mathcal{D}} \text{Acc}(\mathcal{E})$ \Comment{Best accuracy}
    \State $\mathcal{D}^* \gets \{(\mathcal{E}, \cdot) \in \mathcal{D} : \text{Acc}(\mathcal{E}) = a^*\}$ \Comment{Tie candidates}
    \State $\mathcal{E}^*_t \gets \arg\max_{\mathcal{E} \in \mathcal{D}^*} \left( \text{Contrastive}(\mathcal{E}), -|\mathcal{E}| \right)$ \Comment{Lexicographic}
    
    \If{$t < T$}
        \State \textbf{Phase 2: \textcolor{Emerald}{Generation}}
        \For{$(x_i, y_i) \in \mathcal{S}$}
            \State $r_i^{(t)} \gets \textsc{ContrastiveInfill}(x_i, y_i, \mathcal{E}^*_t)$
        \EndFor
        \State $\mathcal{P}_{t+1} \gets \textsc{Merge\&Cap}(\mathcal{P}_t, \{(x_i, y_i, r_i^{(t)})\})$
    \EndIf
\EndFor
\State \Return $\mathcal{E}^*_T$
\end{algorithmic}
\end{algorithm}

\vspace{-19pt}

\paragraph{\bf Optimization Objective}
To operationalize the concept of grading quality, we define a multi-objective function $g(\mathcal{E})$ that balances global accuracy with the structural properties of the context window. While high validation accuracy is the primary signal of success, relying on it alone can lead to unstable results due to the stochastic nature of LLMs. Therefore, we incorporate regularization terms that prioritize sparsity and pedagogical discriminability. Overall, we optimize three objectives simultaneously: (1) validation accuracy $\text{Acc}(\mathcal{E}, \mathcal{V})$, (2) sparsity $-|\mathcal{E}|$ to encourage the selection of concise subsets that reduce token usage and inference latency, and (3) contrastive density $\text{Contrastive}(\mathcal{E})$ to reward boundary pair coverage. 
Rather than manually tuning fixed weights, we adopt \emph{Tchebycheff scalarization} with randomized weights \cite{bowman1976relationship}, which converts the multi-objective problem into a sequence of single-objective problems: 
\vspace{-6pt}
\begin{equation*}
g(\mathcal{E}) = \max\left( w_1 \cdot (\text{Acc}(\mathcal{E}, \mathcal{V}) - \text{Acc}^*), \; w_2 \cdot (-|\mathcal{E}|) \right) + w_3 \cdot \text{Contrastive}(\mathcal{E})
\end{equation*}
where $\text{Acc}^*$ is the best accuracy observed so far, and the weights $(w_1, w_2, w_3)$ are randomly sampled at each BO iteration. This randomization enables exploration of diverse trade-offs on the Pareto front without requiring manual hyperparameter tuning.
The crucial component for enhanced reliability is the $\text{Contrastive}(\mathcal{E})$ term, which quantifies the subset's ability to define decision boundaries. We calculate this score based on the density of ``boundary pairs'' within the selected subset. A pair of exemplars $(e_i, e_j)$ is considered a boundary pair if they possess high semantic similarity yet carry distinct ground truth labels. Formally, this is defined where $\text{sim}(e_i, e_j) \geq \tau$ and $\ell(e_i) \neq \ell(e_j)$ ($\text{sim}(\cdot)$\footnote{Cosine similarity between embeddings from \texttt{text-embedding-3-small}. Embeddings are computed for each exemplar by concatenating the input response and rationale.} denotes cosine similarity and $\ell(\cdot)$ represents the ground truth label). By maximizing this term, the objective function explicitly rewards subsets that present the model with boundary scenarios (i.e., instances where adjacent scores are separated by subtle linguistic or logical nuances) thereby teaching the model to distinguish between grade levels rather than merely recognizing surface patterns.

\vspace{-19pt}

\subsubsection{Candidate Generation via Contrastive Operators}
To effectively navigate the combinatorial search space, we approach candidate generation from two complementary perspectives: greedy exploitation and global exploration. The greedy component focuses on iteratively refining the currently best-performing subset, $\mathcal{E}_{\text{best}}$, by generating slightly different candidates yet with structurally significant modifications. This local refinement is driven by our novel \emph{Contrastive Operators}, which are designed to actively manipulate the decision boundaries within the context window. For each exemplar $e_i \in \mathcal{E}_{\text{best}}$, we first identify a set of boundary candidates $\mathcal{C}_i$ from the broader pool $\mathcal{P}_t$ defined by: 
$\mathcal{C}_i = \{ e_j \in \mathcal{P}_t \setminus \mathcal{E}_{\text{best}} \mid \text{sim}(e_i, e_j) \ge \tau \land |\ell(e_i) - \ell(e_j)| = 1 \}$. 
This constraint ensures that we focus exclusively on candidates that are semantically proximal to existing demonstrations yet belong to a strictly adjacent label.

We utilize this set through two distinct operations that alter the decision space in different ways. First, the \emph{Contrastive-Add} operator inserts a candidate $e_j \in \mathcal{C}_i$ into the existing subset alongside $e_i$. By placing two semantically similar but distinctly labeled examples side-by-side, this operation increases the resolution of the context, forcing the model to attend to the specific micro-features that distinguish the  grades, effectively \emph{sharpening} the decision boundary. However, relying solely on addition would cause the subset to grow monotonically, eventually exceeding the context window or inflating inference costs. To mitigate this, we employ the \emph{Contrastive-Swap} operator, which replaces $e_i$ with $e_j$. This operation \emph{recalibrates} the location of the decision boundary within the semantic space, allowing the optimizer to test whether shifting the local standard for a specific concept improves global accuracy. We also allow standard \emph{One-Flip} perturbations to further fine-tune the subset \cite{johnson1988easy}. Conversely, the exploration component prevents the algorithm from stagnating in local optima through \emph{Random Sampling}, which introduces subsets drawn independently from the entire pool $\mathcal{P}_t$. Finally, to ensure the search is continually anchored by high-quality pedagogical priors, the manually curated expert subset is invariably added to the candidate pool in every iteration.

\vspace{-12pt}
\subsubsection{Efficient Evaluation and Selection}
To select the best candidate from the generated pool without running expensive inferences on all of them, we utilize a Gaussian process (GP) surrogate. Since exemplar subsets are discrete combinatorial objects, we encode them as binary membership vectors and project them into a continuous latent space. The GP is fitted on historically observed pairs of (subset, objective value), allowing it to predict the mean quality $\mu$ and uncertainty $\sigma$ for all new candidates. We employ the Expected Improvement (EI) acquisition function to select the most promising candidate for actual LLM evaluation: $
    \text{EI}(\mathcal{E}) = (\mu(\mathcal{E}) - g^+) \Phi(z) + \sigma(\mathcal{E})\phi(z)$, where $g^+$ represents the best objective value observed so far, and $\Phi$, $\phi$ are the standard normal CDF and PDF. The EI criterion naturally balances the exploitation of high-performing regions with the exploration of uncertain configurations. Once the chosen candidate is evaluated via the LLM, the result is used to update the GP, and the cycle repeats (Lines 10--15). Upon the exhaustion of the evaluation budget, the final subset is selected using a lexicographic criterion (Lines 16--18) that prioritizes validation accuracy first, followed by the contrastive score, ensuring that the final output is not only accurate but pedagogically robust.

\subsection{Phase 2: Discriminative Rationale Generation}
\label{sec:phase2}

\begin{figure}[htbp]
 \vspace{-0.36in}
    \centering
    \includegraphics[width=\linewidth]{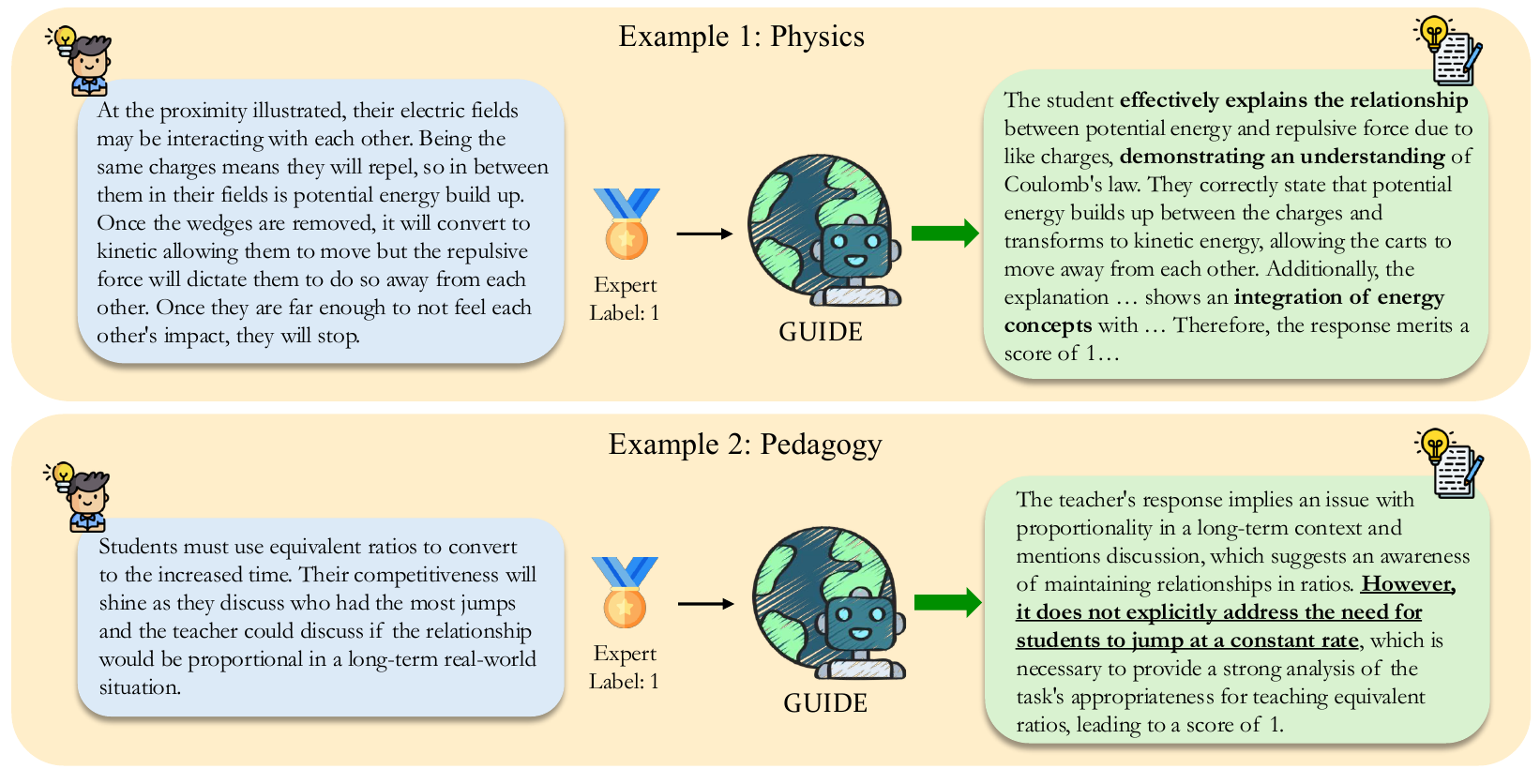}
    \vspace{-0.2in}
    \caption{Illustrative examples of demonstrations (student response, expert label, and synthetic rationale) generated by GUIDE.} 
    \label{fig:guide_examples}
    \vspace{-0.2in}
\end{figure}

The second phase creates a feedback loop to improve the quality of the candidate pool $\mathcal{P}_t$ itself. The logic driving this iteration is that the quality of a generated rationale is strictly dependent on the quality of the context provided during its generation. The optimal subset $\mathcal{E}^*_t$ identified in Phase 1 is, by definition, highly effective at distinguishing grade boundaries. We utilize this subset to regenerate rationales for the training data $\mathcal{S}$, thereby creating higher-quality candidates for subsequent optimization rounds (Lines 21--24).

\paragraph{\bf Discriminative Rationale Generation}
GUIDE employs \emph{discriminative rationale generation} via \textbf{contrastive infill}. For each training example $(x_i, y_i) \in \mathcal{S}$, we perform teacher-forced infilling where the LLM is prompted to generate the rationale $r_i$ conditioned on the correct label $y_i$ and the optimal exemplar set $\mathcal{E}^*_t$ as context. As shown in Fig.~\ref{fig:infill_prompt}, the prompt explicitly requires the model to distinguish the true score from adjacent grades. This forces the articulation of specific rubric constraints and produces the high-quality semantic vectors necessary for contrastive operators.

\paragraph{\bf Size Regulation}
After each round $t$, the generated rationales form a new batch of exemplars that are merged into the existing pool:
\vspace{-0.1in}
\begin{equation*}
    \mathcal{P}_{t+1} \leftarrow \mathcal{P}_t \cup \{(x_i, y_i, r_i^{(t)}) : (x_i, y_i) \in \mathcal{S}\}
\end{equation*}

New exemplars are merged via deduplication, where two exemplars are considered duplicates only if they share identical input, label, and rationale triplets after normalization. This allows the pool to contain multiple rationale variants for the same input-label pair, enabling the optimizer to discover the most effective phrasing. To maintain tractable optimization, we enforce a maximum pool size $N_{\max}$. If $|\mathcal{P}_{t+1}| > N_{\max}$, we apply a reduction procedure (Line 24) that partitions the pool into protected expert exemplars $\mathcal{P}_{\text{exp}}$ and generated exemplars $\mathcal{P}_{\text{gen}}$. We retain all expert exemplars and sample $N_{\max} - |\mathcal{P}_{\text{exp}}|$ generated exemplars uniformly at random. This ensures that pedagogical priors are never discarded while maintaining diversity among the generated candidates.

\vspace{-0.2in}
\begin{figure}[h]
    \centering
    \fbox{
    \begin{minipage}{0.78\linewidth}
        \scriptsize
        \textbf{Instruction for Score 0:}\\
        \textit{``Explain why this deserves a 0 (not a 1 or 2). Specifically mention what is \textsc{missing} that would be needed for a higher score.''}
        
        \vspace{0.5em}
        \hrule
        \vspace{0.5em}

        \textbf{Instruction for Score 1:}\\
        \textit{``Explain why this deserves a 1 (not a 0 or 2). Mention what \textsc{prevents} it from being a 0, and what is \textsc{missing} for a 2."}
        
        \vspace{0.5em}
        \hrule
        \vspace{0.5em}

        \textbf{Instruction for Score 2:}\\
        \textit{``Explain why this deserves a 2 (not a 0 or 1). Specifically mention what makes this \textsc{sufficient} for the highest score.''}
    \end{minipage}
    }
    \caption{Sample prompts for discriminative rationale generation. The model is prompted to articulate specific boundary conditions based on the label.}
    \vspace{-0.3in}
    \label{fig:infill_prompt}
\end{figure}

\subsection{Inference}
\vspace{-5pt}

After $T$ rounds of optimization, the framework converges on the final optimal demonstration set $\mathcal{E}^*_T$. During inference, this frozen set serves as the static in-context prompt for all incoming test instances: 
\begin{equation*}
    \hat{y} = \operatorname*{argmax}_{y \in \mathcal{Y}} P_{\mathcal{M}}(y \mid x_{\text{test}}, \mathcal{I}, \mathcal{R}, \mathcal{E}^*_T)
\end{equation*}
Unlike dynamic retrieval methods that select different examples for each query, GUIDE relies on the optimized $\mathcal{E}^*_T$ to provide a generalized decision surface. 
This ensures that the discriminative criteria learned during optimization are consistently applied to grading new submissions.

\section{Experiments}
\label{sec:experiments}

\vspace{-10pt}

This section evaluates whether \textsc{GUIDE} improves automated grading. Our primary focus is on rubric \emph{boundaries} where a student response falls between two adjacent score levels. These are the most difficult cases to grade and are most likely to be misgraded by exactly one point. 

\vspace{-12pt}

\subsection{Experimental Setup}
\label{sec:exp_setup}
\vspace{-4pt}
\subsubsection{Datasets}
\label{sec:datasets}
We conduct experiments on three datasets that represent different grading contexts in science education and teacher education. The first dataset $\mathcal{D}_I$ contains 314 high school student responses to one assessment item associated with 11 rubrics. This dataset tests students' knowledge of electrical interactions in high school physical science \cite{kaldaras2021developing}. $\mathcal{D}_I$ is assessed with binary labels $\{0,1\}$, where 0 denotes requirement `not met' and 1 denotes requirement being `met' for each of the 11 ideas targeted in the rubric. As $\mathcal{D}_I$ lacks expert rationales, GUIDE initializes by synthesizing rationales from scratch. 
The second dataset $\mathcal{D}_C$ is a chemistry education dataset developed within the 3DLP framework \cite{he2023predicting}. $\mathcal{D}_C$ contains two tasks, each assessed along two dimensions: DCI (disciplinary core ideas) and SEP (science and engineering practices). $\mathcal{D}_C$ is graded with ordinal labels $\{0,1,2\}$, where 0 denotes the requirement is `not met', 1 denotes requirement is `partially met', and 2 denotes requirement is `fully met'. 
The third dataset $\mathcal{D}_T$ is a more complex teacher education dataset, which assesses teacher's nuanced interpretations regarding pedagogical knowledge in teaching mathematics \cite{copur2022mathematics}. $\mathcal{D}_T$ contains four grading tasks, each using ordinal labels $\{0,1,2\}$. Overall, we divide each dataset into train, validation, and test sets using a ratio of 3:1:1. The full statistical details of these datasets are in Table \ref{tab:dataset_stats}. 

\vspace{-24pt}
\begin{table}[ht]
\centering
\caption{A summary of statistics of all datasets.}
\label{tab:dataset_stats}
\setlength{\tabcolsep}{10pt}
\resizebox{\textwidth}{!}{%
\begin{tabular}{ccccc}
\hline
\textbf{Dataset} & \textbf{Domain} & \textbf{\#Samples} & \textbf{Scores} & \textbf{\#Expert Examples} \\
\hline
$\mathcal{D}_I$ & Interaction & 314 & \{0,1\} & 0 \\
$\mathcal{D}_T$ & Teacher Education & 229$\sim$236 & \{0,1,2\} & 3$\sim$5 per category \\
$\mathcal{D}_C$ & Chemistry & 163$\sim$184 & \{0,1,2\} & 3 per category  \\
\hline
\end{tabular}
}
\end{table}

\vspace{-34pt}

\subsubsection{Model and Prompt}

We utilize \texttt{GPT-4o-mini}~\footnote{\url{https://platform.openai.com/docs/models/gpt-4o-mini}} as the backbone for all grading, rationale generation, and inference tasks. 
To support the contrastive selection mechanism described in Section \ref{sec:methodology}, we employ \texttt{text-embedding-3-small}~\footnote{\url{https://platform.openai.com/docs/models/text-embedding-3-small}} to compute the pairwise similarity matrix required for the contrastive operators and the $\text{Contrastive}(\mathcal{E})$ objective.

The optimization process is configured to run for $T=5$ rounds. During Bayesian optimization in each round, the Gaussian process evaluates $N_{eval}=32$ candidate subsets. To maintain a balance between context richness and a concise demonstration subset, we constrain the demonstration set size $|E|$ to $[4, 16]$. The semantic similarity threshold for constituting a boundary pair $\tau=0.7$. The high threshold ensures that the $\text{Contrastive}(\mathcal{E})$ objective only rewards pairs that are semantically very close yet possess divergent labels. The candidate pool for the acquisition function is set to 256, and the maximum size of global exemplar pool $\mathcal{P}$ is capped at $N_{\max}=512$ to ensure computational tractability.  
For GUIDE experiments, we use Tchebycheff scalarization with weights sampled at each iteration: $w_1 \sim \text{Uniform}(0.25, 1.0)$, $w_2 = 0.8 \cdot (1 - w_1)$, and $w_3 = 0.2 \cdot (1 - w_1)$. This eliminates the need for manual weight tuning while enabling exploration of diverse accuracy-parsimony-contrastive trade-offs.
The temperature is set to 0.2 for all generation and evaluation calls to ensure stability while allowing for slight variations.

All inputs are formatted using a fixed prompt template across all experiments. The prompt structure begins with system instruction and full rubric, followed by the selected in-context demonstrations, and ends with the target student response. Each demonstration contains student response, expert-assigned score, and an expert or generated rationale. All methods use the same task instruction and rubric and only differ in the selection of in-context demonstrations. 

\vspace{-12pt}
\subsubsection{Baseline Methods}
We compare our framework with five representative exemplar selection baselines, including static strategies (fixed sets for all queries), dynamic strategies (query-specific retrieval), and optimization-based methods. All baselines use the same initial candidate pool and underlying LLM as our main method. The baseline details are as follows: (a) \textbf{\textsc{Naive}} uses a fixed demonstration set without optimization or selection beyond directly including the provided exemplars. This represents the standard ``few-shot'' or ``zero-shot'' performance. (b) \textbf{\textsc{Random}} randomly samples a fixed set of $k$ exemplars from the training pool, keeping the subset constant for all test queries. This verifies whether sophisticated selection strategies outperforms chance selection. (c)\textbf{\textsc{KNN SBERT}} \emph{dynamically} retrieves the top-$k$ most semantically similar exemplars for each query using \texttt{all-MiniLM-L6-v2} (SBERT) \cite{reimers2019sentence} and cosine similarity. This tests the hypothesis that semantic similarity alone yields the most relevant rubric context for grading specific inputs. (d)\textbf{\textsc{Vote-$k$}} constructs a diverse static set of $k$ exemplars by maximizing the minimum distance between selected examples, ensuring broad coverage of the semantic space rather than clustering around the query. (e) \textbf{\textsc{BRIDGE}}~\cite{wan2025few} uses an optimize-generate loop to find demonstrations that maximizes general validation performance. Unlike \textsc{GUIDE}, it does not explicitly target rubric boundaries or generate discriminative rationales.

\begin{table*}[t]
\vspace{-0.2in}
\centering
\footnotesize
\setlength{\tabcolsep}{3.2pt}
\caption{Dataset-level averages (arithmetic mean over categories/tasks).}
\label{tab:chem_te_inte_results}
\resizebox{\textwidth}{!}{%
\begin{tabular}{lccc cccc cccc}
\toprule
& \multicolumn{3}{c}{$\mathcal{D}_I$ (11 categories)} 
& \multicolumn{4}{c}{$\mathcal{D}_C$ (4 tasks)} 
& \multicolumn{4}{c}{$\mathcal{D}_T$ (4 tasks)} \\
\cmidrule(lr){2-4}\cmidrule(lr){5-8}\cmidrule(lr){9-12}
Method
& Acc & QWK & AdjErr
& Acc & QWK & AdjErr & NonAdjErr
& Acc & QWK & AdjErr & NonAdjErr \\
\midrule
Random     & 0.75 & 0.43 & 0.25 & 0.58 & 0.32 & 0.40 & \second{0.03} & 0.59 & 0.54 & 0.38 & \third{0.03} \\
KNN SBERT  & \third{0.78} & 0.44 & \third{0.21} & 0.58 & 0.26 & 0.40 & \third{0.05} & 0.52 & 0.52 & 0.42 & \second{0.02} \\
Vote-$K$   & 0.73 & \third{0.45} & 0.27 & 0.62 & 0.38 & 0.36 & \second{0.03} & \third{0.60} & \third{0.58} & 0.38 & \best{0.01} \\
Naive      & 0.74 & 0.42 & 0.26 & \third{0.69} & \third{0.39} & \third{0.31} & \best{0.00} & 0.59 & 0.54 & \third{0.37} & 0.04 \\
BRIDGE     & \second{0.90} & \second{0.57} & \second{0.19} & \second{0.76} & \second{0.53} & \second{0.24} & \best{0.00} & \second{0.66} & \second{0.65} & \second{0.32} & \second{0.02} \\
GUIDE      & \best{0.92} & \best{0.62} & \best{0.08} & \best{0.80} & \best{0.59} & \best{0.20} & \best{0.00} & \best{0.71} & \best{0.67} & \best{0.28} & \second{0.02} \\
\bottomrule
\end{tabular}%
}
\vspace{2pt}
{\scriptsize \textit{Notes:} \best{red} / \second{green} / \third{blue} denote the best / second-best / third-best method for each metric.}
\vspace{-0.2in}
\end{table*}

\vspace{-12pt}
\subsubsection{Metrics}
To ensure a comprehensive evaluation, we use three specific metrics. First, accuracy (exact match) measures the percentage of times the model predicts the exact correct score. 
Second, the quadratic weighted kappa (QWK) is particularly relevant for ordinal scales (like 0, 1, 2) because it penalizes disagreements based on their distance. In this metric, confusing a 0 with a 2 is penalized much more heavily than confusing a 0 with a 1. Thirdly, to directly diagnose boundary failures, we compute confusion matrix-based error rates.
For ordinal labels $\{0,1,2\}$ ($\mathcal{D}_{T}$ and $\mathcal{D}_{C}$), we define the adjacent error rate (AdjErr) as the function of predictions that are off by exactly one level (e.g., instances where the model predicts a 1 when the truth is 2). 
The non-adjacent error rate (NonAdjErr) measures predictions that skip a level, such as predicting 0 when the truth is 2, which represents a severe failure of logic. 
For the binary dataset $\mathcal{D}_{I}$, \text{NonAdjErr} is always zero by definition, and AdjErr captures all misclassifications.

\vspace{-8pt}

\subsection{Results}
\vspace{-5pt}
\label{sec:exp_results}

\subsubsection{Overall grading quality}
Table~\ref{tab:chem_te_inte_results} presents the aggregate performance across all datasets. On the ordinal datasets $\mathcal{D}_C$ and $\mathcal{D}_T$, \textsc{GUIDE} consistently outperforms the baselines. As shown in the dataset-level averages, \textsc{GUIDE} achieves the highest accuracy and QWK across the board. On $\mathcal{D}_T$, \textsc{Naive} baseline struggles with an accuracy of $0.59$ and QWK of $0.54$, while \textsc{GUIDE} achieves substantial gains with an accuracy of $0.71$ and QWK of $0.67$. This represents a relative accuracy improvement of approximately $20\%$ over the naive baseline. On $\mathcal{D}_C$, \textsc{GUIDE} similarly dominates with an accuracy of $0.80$ and QWK of $0.59$, compared to the \textsc{Naive} accuracy of $0.69$ and QWK of $0.39$. On the interaction dataset $\mathcal{D}_I$, \textsc{GUIDE} demonstrates robust generalization to authentic student language, achieving the highest accuracy of $0.92$ and QWK of $0.62$, significantly surpassing standard baselines like \textsc{Random} (accuracy=$0.75$) and \textsc{Naive} (accuracy=$0.74$).

\vspace{-12pt}

\subsubsection{Boundary error decomposition}
A central hypothesis of \textsc{GUIDE} is that boundary-focused exemplars reduce confusion between adjacent scores. The error decomposition in Table \ref{tab:chem_te_inte_results} confirms this. Across all three datasets, \textsc{GUIDE} yields the lowest AdjErr among all methods. On $\mathcal{D}_I$, \textsc{GUIDE} reduces AdjErr to $0.08$, a massive reduction compared to $0.26$ for \textsc{Naive}. On $\mathcal{D}_T$ and $\mathcal{D}_C$, \textsc{GUIDE} similarly achieves the lowest adjacent error rates of $0.28$ and $0.20$ respectively. Crucially, this reduction in AdjErr does not come at the cost of increased gross errors. The NonAdjErr remains negligible ($0.00$ and $0.02$) for \textsc{GUIDE}, indicating that the primary challenge in these datasets is indeed boundary disambiguation, which the method effectively addresses.

\vspace{-15pt}

\subsubsection{Performance comparison vs. BRIDGE}
Comparing \textsc{GUIDE} directly to \textsc{BRIDGE} highlights the value of boundary-targeted optimization over general metric optimization. While \textsc{BRIDGE} consistently improves over \textsc{Naive}, \textsc{GUIDE} achieves further gains across all metrics. For $\mathcal{D}_T$, \textsc{BRIDGE} reaches an accuracy of $0.66$, whereas \textsc{GUIDE} advances this to $0.71$. Similarly, on $\mathcal{D}_C$, \textsc{GUIDE} outperforms \textsc{BRIDGE} with an accuracy of $0.80$ compared to $0.76$. The distinction is also evident in boundary errors; on $\mathcal{D}_I$, \textsc{GUIDE} achieves an AdjErr of $0.08$ compared to $0.19$ for \textsc{BRIDGE}. This suggests that explicitly contrasting adjacent labels and generating discriminative rationales provides a stronger learning signal for the LLM than simply optimizing for validation accuracy.

\section{Discussion}
\vspace{-4pt}

\subsubsection{The Importance of Boundary-Focused Learning}
Our findings suggest that one key challenge in automated grading lies in distinguishing between adjacent scores. Standard retrieval methods prioritize semantic similarity, which retrieve keyword-heavy examples that fail to clarify the grading criteria. By explicitly optimizing for ``contrastive density,'' GUIDE forces the model to confront ``boundary pairs'' that look similar but earn different scores. Experimental results demonstrate that focusing on these specific edge cases significantly sharpens its ability to apply the rubric correctly. This implies that effective ICL for grading requires examples that act as counterpoints rather than just representatives.

\vspace{-12pt}

\subsubsection{The Role of Discriminative Rationales}
A critical component of GUIDE is the generation of discriminative rationales. Our ablation studies reveal that standard expert rationales are not always sufficient for disambiguation. The effectiveness of GUIDE lies in its ``boundary'' reasoning capability. By prompting the model to explain not only why a score is appropriate but also why adjacent scores are inappropriate, we provide the grader with a clearer decision surface. 

\vspace{-12pt}

\subsubsection{Efficiency and Scalability in Educational Contexts}
From a practical perspective, GUIDE offers a solution to the ``cold-start'' problem in automated grading bypassing the thousands of labeled examples required by traditional supervised learning. GUIDE can achieve high reliability with a small, optimized context window of 4 to 16 examples. Furthermore, the framework's ability to synthesize rationales releases educators from manually writing detailed feedback for every training example. This allows the system to bootstrap from a small set of grades, making high-quality automated assessment accessible for new courses or changing curricula with minimal human effort.

\vspace{-12pt}

\subsubsection{Budget}
Although optimization relies on an iterative loop with multiple LLM calls during training, the financial cost remains highly accessible. Utilizing the cost-effective \texttt{GPT-4o-mini} model minimizes the optimization budget, with single-item optimization typically costing \$5-8. This represents a manageable one-time investment, as the resulting optimized prompt can be deployed indefinitely for inference at a minimal marginal cost. 

\vspace{-12pt}

\subsubsection{Limitations and Future Directions}
Future work should explore this boundary-focused optimization to multimodal tasks, such as grading diagrams or mathematical derivations, where the definition of a semantic neighbor may be complex.

\vspace{-0.1in}
\section{Conclusion}
\vspace{-0.1in}
In this work, we addressed the reliability in LLM-based automated grading by shifting the focus from general accuracy to boundary precision with GUIDE. This framework iteratively selects and refines in-context demonstrations using novel contrastive operators, targeting semantically similar examples with different grades to clarify decision boundaries.
Experiments across physics, chemistry, and teacher education datasets demonstrate that 
GUIDE significantly reduces adjacent grading errors and shows potential for high-quality, human-aligned automated assessment.


%
%
%
\bibliographystyle{splncs04}
\bibliography{secs/ref}
%




\end{document}